\documentclass[lettersize,journal]{IEEEtran}
\usepackage{amsmath,amsfonts}
\usepackage{algorithmic}
\usepackage{algorithm}
\usepackage{array}
\usepackage[caption=false,font=normalsize,labelfont=sf,textfont=sf]{subfig}
\usepackage{textcomp}
\usepackage{stfloats}
\usepackage{url}
\usepackage{verbatim}
\usepackage{graphicx}
\usepackage{cite}
\usepackage{makecell}
\usepackage{booktabs}
\usepackage{tabularx}
\usepackage{array}
\usepackage{multirow}

\begin{document}

\title{FLM: Frequency-Aware Language Models for Generative Image Compression}

\author{
  Jiarun~Chen,
  Kejun~Wu,~\IEEEmembership{Senior Member,~IEEE},
  Li~Li,~\IEEEmembership{Senior Member,~IEEE},
  Chengtao~Cai,~\IEEEmembership{Senior Member,~IEEE},
  Zhengguo~Li,~\IEEEmembership{Fellow,~IEEE},
  and Chia-Wen~Lin,~\IEEEmembership{Fellow,~IEEE}
  \thanks{Jiarun~Chen and Kejun~Wu are with the School of Electronic Information and Communications, Huazhong University of Science and Technology, Wuhan 430074, China.}
  \thanks{Li~Li is with the MOE Key Laboratory of Brain-Inspired Intelligent Perception and Cognition, University of Science and Technology of China, Hefei 230093, China}
  \thanks{Chengtao~Cai is with College of Intelligent Systems Science and Engineering, Harbin Engineering University, Harbin 150001, China.}  
  \thanks{Zhengguo~Li is with VI Department, Institute for Infocomm Research, Agency for Science, Technology and Research (A*STAR), Singapore.}
  \thanks{Chia-Wen Lin is with the Department of Electrical Engineering and the Institute of Communications Engineering, National Tsing Hua University, Hsinchu 30013, Taiwan.}
  \thanks{This work was supported by the National Natural Science Foundation of China under Grant 62501246.}
  \thanks{Corresponding author: Kejun~Wu (kjwu@hust.edu.cn).}
  }

\markboth{Journal of \LaTeX\ Class Files,~Vol.~14, No.~8, August~2026}%
{Shell \MakeLowercase{\textit{et al.}}: A Sample Article Using IEEEtran.cls for IEEE Journals}

\maketitle
\begin{abstract}

Generative models have significantly improved the performance ceiling of image lossy compression at low bitrates by exploiting learned priors.
However, the generated textures and semantic details may deviate from the source content, thereby affecting the fidelity of image reconstruction.
To solve these challenges, we propose FLM, a frequency-aware language model that improves compression efficiency through frequency-domain probabilistic modeling while retaining deterministic reconstruction. At the encoder, the input image is transformed into quantized DCT coefficients, which are organized into discrete sequences using macroblock-based coefficient tokenization. FLM then performs next-coefficient prediction to autoregressively estimate token-wise conditional probability distributions for arithmetic coding, thereby generating a compact bitstream. 
At the decoder, the LLM and arithmetic decoder jointly recover the frequency-domain data, followed by inverse transformations for image reconstruction. 
A task-specific frequency-domain dataset and a two-stage fine-tuning strategy are further developed to enable the model to operate across multiple bitrate settings. 
FLM is a versatile compressor that is compatible with both lossy compression and lossless JPEG recompression frameworks. 
Experiments show that FLM exceeds conventional and generative lossy compression methods in rate-distortion performance.
FLM achieves BD-PSNR gains of 3.30 dB, 3.83 dB, and 3.80 dB than JPEG baseline on Kodak, Tecnick, and CLIC2020, respectively.
Better qualitative quality of FLM can be achieved in improving semantically high fidelity and suppressing blocking artifacts.
FLM is also validated to be applicable to the lossless recompression task with competitive performance.  
\end{abstract}


\section{Introduction}
\begin{figure}[t]
    \centering
    \includegraphics[width=\columnwidth]{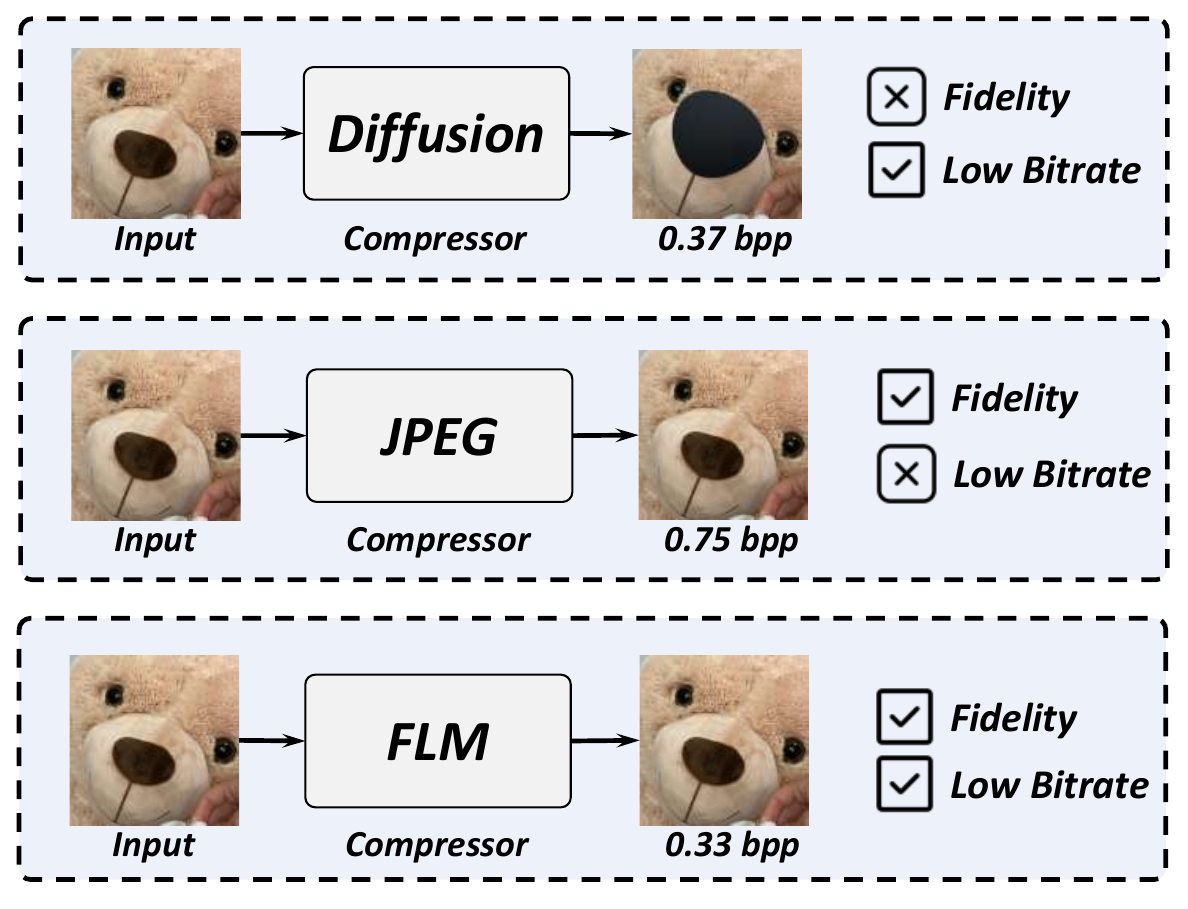}
    \caption{Conceptual comparison of representative image compression methods. Diffusion-based methods compromise reconstruction fidelity, whereas JPEG requires a higher bitrate. Our FLM method achieves both high fidelity and low bitrate.}
    \label{fig:FIG1}
\end{figure}

\IEEEPARstart{I}{mage} compression seeks to represent visual signals with as few bits as possible while preserving the information required for faithful reconstruction. Recent advances in generative models such as generative adversarial networks (GANs)~\cite{goodfellow2014generative} and diffusion models~\cite{ho2020denoising} have widely used in image generation, and digital forensics~\cite{10812851}.
By exploiting learned image priors, generative models also have established as an important paradigm for image compression \cite{10855472,relic2024lossy}. 
Generative image compression can reconstruct perceptually compelling images from highly compact representations, particularly at low bitrates. However, perceptual realism does not necessarily imply fidelity to the source image \cite{yan2021perceptual,yan2022optimally}, as shown in Fig.~\ref{fig:FIG1}. Direct synthesis of image content from an information-limited representation may yield textures, structures, and semantic details that appear plausible yet differ from those in the source image. Consequently, improving compression efficiency without introducing content deviations associated with generative synthesis remains an important challenge.

Conventional Transform-based codecs follow a different reconstruction principle. Methods such as JPEG \cite{wallace1991jpeg} and JPEG-XL \cite{alakuijala2019jpeg} represent and reconstruct an input image using explicitly derived frequency-domain coefficients. Although transform and quantization operations still introduce distortion, this reconstruction process does not introduce the type of content deviation caused by generative image synthesis. Nevertheless, conventional entropy models exploit only limited contextual dependencies and may not fully characterize the complex spatial, cross-frequency, and long-range relationships among frequency-domain coefficients, leaving considerable statistical redundancy unexploited \cite{sun2022lossless,xiang2026efficient,9381234}.

Beyond direct content synthesis, generative models can also support compression through probability estimation. Delétang et al. \cite{deletang2024language} demonstrated the close relationship between autoregressive language modeling and lossless compression by combining model-predicted symbol probabilities with arithmetic coding \cite{witten1987arithmetic}. Subsequent studies adapted large language models (LLMs) to the lossless compression of image pixels or prediction residuals \cite{li2025lossless,chen2026large,du2025large,zheng2025joint}. These methods establish the feasibility of employing LLMs as strong entropy models.

However, existing LLM-based image compression methods primarily perform lossless coding of pixels or pixel residuals. These methods require a high bitrate to encode intrinsic image noise and subtle textures that are barely perceptible to the human eye. In contrast, frequency-domain representations explicitly separate frequency components and allow them to be treated according to their perceptual importance. Transform and quantization drive many high-frequency coefficients to or near zero, thereby yielding sparse and structured sequences that are easier for the LLM to model and predict. However, using an LLM to model frequency-domain coefficients presents several challenges. For example, frequency-domain coefficients have a wide dynamic range that varies with quantization strength. Moreover, direct-current and alternating-current coefficients exhibit different numerical ranges and statistical distributions.

To address these challenges, we propose FLM, a frequency-aware language model for generative image compression, which extends the LLM-based compression paradigm from pixel-domain coding to frequency-domain probabilistic modeling. Specifically, at the encoder, FLM first applies the JPEG processing pipeline to the input image to obtain quantized DCT coefficients. A macroblock-based DCT coefficient tokenization strategy then organizes spatially adjacent coefficient blocks into discrete token sequences, which are subsequently fed into FLM for autoregressive estimation of the conditional probability distribution of each coefficient. We further propose a range-constrained probability modeling strategy, which improves coding efficiency by restricting the prediction space to coefficient candidates valid for the current macroblock and coefficient type. At the decoder, the arithmetic decoder and the LLM jointly perform autoregressive decoding to progressively recover the frequency-domain sequence from the compressed bitstream. The reconstructed frequency-domain data are subsequently converted back into image pixels through a series of inverse processing steps, yielding the reconstructed image. To adapt a pretrained LLM to the numerical and statistical characteristics of frequency-domain coefficients, we construct a task-specific frequency-domain dataset and develop a two-stage fine-tuning strategy.

The main contributions of this paper are summarized as follows:
\begin{itemize}
    \item We propose FLM, a novel image compression framework that combines conventional DCT-based transform coding with the powerful sequence-modeling capabilities of LLMs. By formulating entropy modeling as a next-coefficient prediction task, FLM effectively leverages the contextual modeling capabilities of LLMs to achieve superior compression efficiency. To the best of our knowledge, FLM is the first LLM-based image compression framework to directly model the conditional distributions of quantized DCT coefficient tokens.
    \item We develop a macroblock-based DCT coefficient tokenization and a range-constrained probability modeling strategy to enable the LLM to better understand frequency-domain coefficients. A task-specific fine-tuning dataset and a two-stage fine-tuning strategy are developed to further enhance FLM’s ability to model coefficient distributions across multiple bitrates.
    \item We conduct comprehensive evaluations on the Kodak, Tecnick, and CLIC2020 datasets against conventional codecs and generative compression methods. The results demonstrate that FLM achieves overall superior rate-distortion performance and qualitative quality against traditional codecs and generative codecs. FLM also validated to obtain competitive performance in lossless JPEG recompression task.
\end{itemize}

\section{Related Work}
\subsection{GAN-based image compression}
Generative adversarial networks (GANs) were introduced by Goodfellow \cite{goodfellow2014generative} in 2014. A GAN primarily consists of two components: a generator and a discriminator. The generator learns to produce samples that approximate the distribution of real images, whereas the discriminator learns to distinguish generated samples from real ones. These two networks are jointly optimized through adversarial training, ultimately yielding a generator capable of producing images that closely resemble real images.

Agustsson et al. \cite{agustsson2019generative} were among the first researchers to apply GANs to image compression task. Their algorithm incorporated a discriminator into the compression framework to guide the training of the neural networks, thereby improving the perceptual quality and fine-detail reconstruction of the decoded images. However, the quality of their reconstructed images was unsatisfactory. Building on this idea, Mentzer et al. \cite{mentzer2020high} improved the reconstruction quality by designing an encoder and a decoder with more sophisticated network architectures. Through theoretical analysis, Yan et al. \cite{yan2021perceptual} presented several important insights. For example, at a given bitrate, the mean squared error required to achieve perfect perceptual quality is twice the minimum achievable mean squared error. Based on these insights, they proposed a training framework that balances rate, distortion, and perception. In their subsequent work, Yan et al. \cite{yan2022optimally} further improved the flexibility of the method by enabling an adjustable trade-off between distortion and perception in the reconstructed images, which could not be achieved by previous methods.

\subsection{Diffusion-based image compression}
In 2020, the introduction of Denoising Diffusion Probabilistic Models (DDPMs) \cite{ho2020denoising} attracted widespread attention to diffusion-based generative models. A diffusion model typically comprises a forward diffusion process and a reverse denoising process. During the forward process, Gaussian noise is progressively added to an image until its distribution approaches a standard Gaussian distribution. During the reverse process, a neural network predicts the noise component at each timestep and progressively estimates the preceding state. Through iterative denoising, the model gradually transforms random Gaussian noise into a generated image \cite{10420512}.

Relic et al. \cite{relic2024lossy} proposed an end-to-end image compression framework based on diffusion models. At the encoder, this framework encodes the latent representation of an image in the semantic space of a diffusion model and compresses it through operations such as quantization and entropy coding. A parameter estimation module is further introduced to determine the quantization level and the number of denoising steps. At the decoder, the latent representation is recovered through entropy decoding and dequantization and is subsequently fed into the diffusion model to reconstruct the image. Lei et al. \cite{lei2023text+} introduced a cross-modal compression method that employs text and contour maps as control signals for the diffusion model. Bordin et al. \cite{bordin2024linearly} proposed using text and color maps as semantic information to guide diffusion-based image generation. Zhang et al. \cite{zhang2025stablecodec} improved the computational efficiency of the compression framework by performing single-step diffusion. They further developed a dual-branch compression framework in which an auxiliary codec serves as an additional branch to enhance the reconstruction of image details. Chen et al. \cite{chen2025efficient} converted high-resolution images into low-resolution representations for encoding. At the decoder, an upsampling algorithm and a diffusion model are employed to restore the spatial resolution and reconstruct image details, respectively, thereby achieving a favorable balance between bitrate consumption and reconstruction quality.

\subsection{LLM-based image compression}
LLM-based image compression was first explored by Delétang et al.\cite{deletang2024language}. Their work investigated the use of an LLM for pixel prediction and supplied the probability distribution predicted for each pixel to an arithmetic encoder for lossless encoding of the source data. Benefiting from the accurate probability estimates produced by the LLM, their method outperformed conventional codecs such as PNG \cite{Thomas1997} and established the fundamental design paradigm for LLM-based image compression.

Building on this framework, subsequent studies introduced a variety of extensions and improvements. Li et al. \cite{li2025lossless} further validated this design paradigm by employing iGPT \cite{chen2020generative}, a large model trained on image data, for pixel prediction. Their method achieved improved compression performance and led to the observation that “better understanding leads to better compression.” Chen et al. \cite{chen2026large} further extended the framework proposed by Delétang et al. and introduced P\textsuperscript{2}-LLM. By modeling semantic dependencies among pixels across color channels, incorporating contextual prompts, and adopting an innovative tokenization strategy, P\textsuperscript{2}-LLM effectively improved compression efficiency. Du et al. \cite{du2025large} proposed an LLM-based lossless image compression method incorporating visual prompts. Specifically, the original image is first subjected to lossy compression, and the pixel-wise residual between the original image and its lossy reconstruction is then computed. The lossy reconstruction is subsequently provided to the LLM as a contextual prompt, supplying semantic priors for residual coding and thereby improving the accuracy of the model in predicting the probability distribution of residual pixels. Furthermore, Zheng et al. \cite{zheng2025joint} extended the application of this framework to medical image compression.

\section{Proposed Method}

\begin{figure*}[t]
    \centering
    \includegraphics[width=1\textwidth]{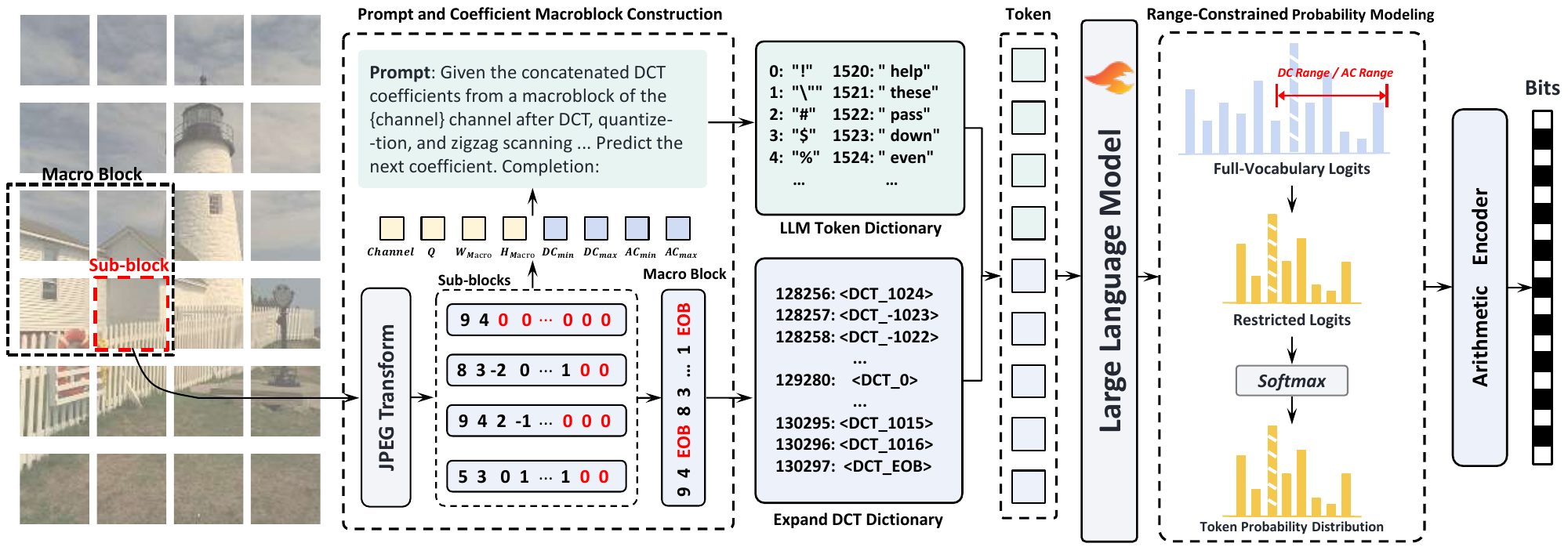}
    \caption{The overall framework of FLM. The input image is converted into quantized DCT coefficients by a JPEG-style transform, and the proposed Macroblock-Based DCT Coefficient Tokenization organizes the coefficients into token sequences. The task-prompt tokens and coefficient tokens are concatenated and fed into the LLM to predict the conditional distribution of the next token. The proposed Range-Constrained Probability Modeling restricts predictions to valid coefficient tokens to improve coding efficiency, and the resulting probability distribution is then used by an arithmetic encoder to generate the bitstream.}
    \label{fig:Main}
\end{figure*}

\subsection{Overall Framework of FLM}
FLM consists of an encoder and a decoder. As illustrated in Fig.~\ref{fig:Main}, at the encoder, the input RGB image is first processed by a JPEG-style transform frontend and converted into DCT coefficients in three color channels. This frontend follows the fundamental transform coding pipeline of JPEG, including color space conversion, chroma subsampling with a default 4:2:0 format, DCT, quantization, and zigzag scanning. Subsequently, FLM applies the proposed macroblock-based DCT coefficient tokenization method to organize the two-dimensional DCT coefficients of each channel into a one-dimensional sequence of discrete tokens. For each token to be encoded, the LLM performs next-coefficient prediction based on the task prompt and previously observed tokens to estimate its conditional probability distribution. The resulting probability distribution is then provided to an arithmetic encoder for entropy coding of the corresponding DCT coefficient token. The final binary bitstream comprises the arithmetic-coded payload and the associated side information, such as the image dimensions and macroblock-level DC and AC coefficient ranges.

The decoding process can be broadly regarded as the inverse of the encoding process. At the decoder side, an LLM with the same architecture and parameter weights as those used by the encoder must be deployed, together with an identical tokenizer and task prompt template. During decoding, the side information is first parsed from the binary bitstream and used to construct the corresponding task prompt. Based on this prompt, the LLM produces the conditional probability distribution of the first DCT coefficient token, which is then used by the arithmetic decoder to recover the corresponding token from the bitstream. The decoded token is subsequently fed back into the LLM as historical context for predicting the conditional probability distribution of the next token. This autoregressive decoding procedure is repeated until the DCT coefficient sequences of all macroblocks have been recovered. Finally, the decoder applies the inverse operations of the transform frontend, including inverse zigzag scanning, dequantization, and inverse DCT. The chroma channels are then upsampled, and the reconstructed RGB image is obtained through inverse color space conversion.
\subsection{Macroblock-based DCT Coefficient Tokenization}
\label{3_2}
\subsubsection{Tokenization Strategy}
Existing large language models generally do not contain dedicated tokens for representing frequency-domain coefficients in their pretrained vocabularies. Therefore, to convert quantized DCT coefficients into discrete sequences suitable for autoregressive modeling by large language models, an appropriate tokenization scheme is required. Delétang \cite{deletang2024language} adopted an index-based direct mapping strategy, in which each subpixel value is mapped to the token whose vocabulary index has the same numerical value. However, vocabulary indices do not inherently carry explicit numerical semantics, making it difficult for this approach to preserve the numerical relationships among different subpixel values. Chen et al. \cite{chen2026large} proposed a numerical-string-based tokenization scheme that represents each subpixel value using a text token with the same numerical meaning. For example, a subpixel value of 15 is mapped to the token corresponding to the string ``1''. Nevertheless, the actual token representation of a numerical string depends on the tokenizer being used. For instance, under LLaMA 3.1 \cite{grattafiori2024llama}, the string ``25'' is encoded as a single token, whereas in the Qwen family of models, it may be split into three tokens, namely ``2'', ``5'', and ``5''. In addition, 8-bit image pixels are non-negative integers within the range [0,255], whereas quantized DCT coefficients have a wider dynamic range and include both positive and negative values. Therefore, directly adopting numerical-string representations may cause some DCT coefficients to be split into multiple tokens, thereby increasing the autoregressive sequence length and introducing additional complexity into sequence management for probability modeling and arithmetic coding.

To address this issue, we extend the pretrained vocabulary of LLaMA 3.2-1B by assigning an independent token to each quantized DCT coefficient within the range [-1024, 1016], such that every coefficient can be represented by a single token. The embedding vectors of the newly introduced tokens are further optimized during subsequent training to learn representations adapted to the distribution of frequency-domain coefficients.

To leverage the numerical representation capability already acquired by the pretrained model, we propose a subword mean-pooling-based embedding initialization method. Let $\mathbf{E}$ denote the input token embedding matrix of the pretrained model, and let
$\mathcal{T}(\operatorname{str}(i))=\left(t_1,t_2,\ldots,t_{m_i}\right)$
denote the subword sequence obtained by tokenizing the numerical string of coefficient $i$ without adding special tokens, where $t_j$ denotes the $j$-th resulting subword token and $m_i$ denotes the total number of subword tokens generated for coefficient $i$. The initial embedding of the newly introduced DCT token is defined as
\begin{equation}
\mathbf{e}_{\left\langle \mathrm{DCT}_{i} \right\rangle}
=
\frac{1}{m_i}
\sum_{j=1}^{m_i}
\mathbf{E}[t_j],
\qquad
i \in [-1024, 1016],
\label{eq:dct_token_initialization}
\end{equation}
When a numerical string is encoded as a single pretrained token, Eq.~\eqref{eq:dct_token_initialization} is equivalent to directly adopting the original embedding of that token. When the numerical string is split into multiple subword tokens, the initial representation of the newly introduced token is obtained by averaging the embeddings of the corresponding subwords. This initialization strategy incorporates the numerical semantics learned by the pretrained model into the newly introduced tokens and provides a more informative starting point for subsequent model fine-tuning than random initialization.

\subsubsection{Macroblock Organization}
This work adopts a JPEG-style transform coding pipeline. Each color channel is partitioned into non-overlapping $8 \times 8$ image blocks, with each block producing 64 frequency-domain coefficients. Independently modeling the 64 coefficients of a single $8 \times 8$ block provides the model with only limited spatial context. To improve the accuracy of the LLM in estimating the probability distribution of frequency-domain coefficients, we treat the coefficient sequence of each $8 \times 8$ image block as a coefficient sub-block. A total of $16 \times 8$ spatially adjacent coefficient sub-blocks are then concatenated in raster-scan order to form a coefficient macroblock, which serves as the basic prediction unit of the large language model.

Considering the pronounced sparsity of quantized DCT coefficients in high-frequency regions, we further introduce a special token $\langle \mathrm{DCT\_EOB} \rangle$ to compactly represent each coefficient sub-block. Specifically, when the AC coefficient sequence obtained through zigzag scanning contains a consecutive run of trailing zeros, the entire zero run is replaced by a single $\langle \mathrm{DCT\_EOB} \rangle$ token. This representation shortens the input sequence while preserving exact reconstruction of the coefficient sequence. The embedding vector of $\langle \mathrm{DCT\_EOB} \rangle$ is initialized using the embedding of the native end-of-sequence token, $\langle \mathrm{EOS} \rangle$, from the pretrained LLaMA vocabulary, thereby incorporating a prior representation associated with sequence-termination semantics.

\begin{figure}[t]
    \centering
    \includegraphics[width=\columnwidth]{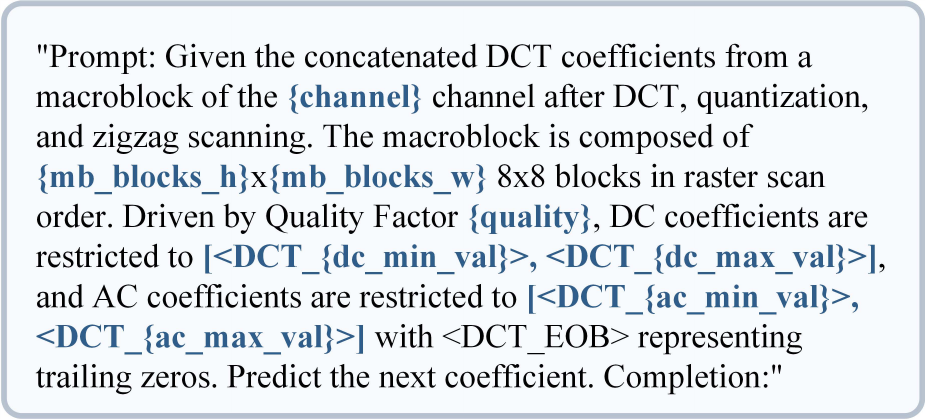}
    \caption{Prompt template used by FLM. For each macroblock, the template is instantiated with the corresponding channel, macroblock dimensions, quality factor, and minimum and maximum DC and AC coefficient values.}
    \label{fig:Prompt Template}
\end{figure}

Motivated by the effectiveness of prompt-based mechanisms in enhancing visual representation learning~\cite{11329502,chen2026large}, we design the prompt template shown in Fig.~\ref{fig:Prompt Template}. This prompt template adopted by FLM incorporates information such as the DCT coefficient processing procedure, channel type, quality factor, DC and AC coefficient ranges, and the coefficient prediction task. The DC and AC coefficient ranges are determined at the encoder by collecting statistics for each coefficient macroblock and are transmitted to the decoder as side information. During both encoding and decoding, the LLM takes the prompt together with the previously known frequency-domain coefficient tokens as contextual input for the prediction of the next-token.

\subsection{Range-Constrained Probability Modeling}
\label{3_3}

FLM controls the trade-off between compression rate and reconstruction quality by adjusting the quantization tables through the quality factor $Q$. A larger $Q$ corresponds to weaker quantization, which generally preserves more frequency-domain information and improves reconstruction quality,  but also results in more complex coefficient distributions and thereby poses greater challenges to frequency-domain coefficient prediction by the LLM. FLM adopts the standard JPEG luminance and chrominance quantization tables and generates the quantization tables associated with different values of $Q$ according to the JPEG quality-factor scaling rule.

Since FLM employs arithmetic coding to entropy-code frequency-domain coefficient tokens, its coding efficiency largely depends on how closely the estimated conditional probability distribution matches the true symbol distribution. When the model assigns a higher and more accurate conditional probability to the target token, the arithmetic coder can represent that token using fewer bits. Note that quantized DCT coefficients exhibit different numerical distributions across quality factors. In particular, at low values of $Q$, stronger quantization maps a large proportion of high-frequency coefficients to zero and confines the remaining nonzero coefficients to a relatively narrow numerical range. If softmax normalization is performed over the entire output vocabulary of the LLM, numerous ordinary tokens that are invalid at the current prediction step, as well as DCT tokens outside the actual coefficient range, are still included in the normalization. These invalid candidates consume part of the probability mass, thereby reducing the normalized probabilities assigned to valid candidate tokens and increasing the expected arithmetic coding length.

To address this issue, we propose a range-constrained probability modeling strategy. This method reuses the macroblock-level DC and AC coefficient ranges collected at the encoder to dynamically restrict the candidate token set at each prediction position. Since DC and AC coefficients differ substantially in both numerical range and statistical distribution, we employ separate macroblock-level ranges to construct their respective candidate sets. At a DC position, the candidate set contains only the DC coefficient tokens within the corresponding range. At an AC position, the candidate set consists of the AC coefficient tokens within the corresponding range together with the special token $\langle\mathrm{DCT\_EOB}\rangle$. 

Given the task prompt $\mathbf{s}$ and the preceding coefficient tokens $\mathbf{y}_{<t}=(y_1,\ldots,y_{t-1})$, the LLM produces a logit $z_{t,v}$ for each token $v$. At prediction position $t$, $\mathcal{C}_t$ denotes the valid candidate-token set. The range-constrained conditional probability is defined as
\begin{equation}
p\!\left(y_t=v\mid\mathbf{s},\mathbf{y}_{<t}\right)
=
\begin{cases}
\frac{\exp(z_{t,v})}{\sum_{u\in\mathcal{C}_t}\exp(z_{t,u})},
& v\in\mathcal{C}_t,\\[6pt]
0, & v\notin\mathcal{C}_t.
\end{cases}
\label{eq:range_constrained_probability}
\end{equation}
Tokens outside $\mathcal{C}_t$ are assigned zero probability, whereas the logits of valid candidates are normalized exclusively over $\mathcal{C}_t$. The resulting conditional probability distribution is then supplied to the arithmetic coder.

\subsection{Two-Stage Fine-tuning Strategy}
To enhance the capability of the LLM to understand and predict frequency-domain coefficients, this work constructs a dedicated frequency-domain coefficient dataset and develops a two-stage fine-tuning strategy.

\subsubsection{Dataset} Following the fine-tuning scheme of Chen et al.~\cite{chen2026large}, we construct the fine-tuning dataset based on DIV2K~\cite{agustsson2017ntire}. The DIV2K training set contains 800 high-quality natural images with 2K resolution and has been widely used to train image super-resolution and learned image compression models. Following the data processing pipeline of FLM, the images are processed by the JPEG-style transform frontend and macroblock-based DCT coefficient tokenization strategy, through which the two-dimensional frequency-domain coefficients are converted into macroblock-level one-dimensional token sequences.

\begin{table}[!htbp]
\centering
\caption{Hyperparameters for two-stage fine-tuning.}
\label{tab:finetuning_settings}
\renewcommand{\arraystretch}{1.12}
\setlength{\tabcolsep}{2pt}

\begin{tabular*}{\columnwidth}{
@{\extracolsep{\fill}}lcc@{}}
\toprule
\textbf{Hyperparameter} &
\textbf{Stage I} &
\textbf{Stage II} \\
\midrule
Training paradigm       & SFT                & SFT             \\
Trainable parameters    & Embedding/output head & LoRA adapters \\
LoRA targets            & --                 & All linear projections \\
LoRA rank               & --                 & 32              \\
\addlinespace[1pt]
Learning rate           & \(5\times10^{-4}\) & \(1\times10^{-4}\) \\
Training epochs         & 1                  & 1               \\
Batch size/GPU          & 1                  & 1               \\
Gradient accumulation   & 8                  & 8               \\
Max. sequence length    & 9000               & 9000            \\
LR scheduler            & Cosine             & Cosine          \\
Warmup ratio            & 0.1                & 0.1             \\
\bottomrule
\end{tabular*}

\vspace{2pt}
\parbox{\columnwidth}{\footnotesize
\textit{Note:} All linear projections include
\texttt{q\_proj}, \texttt{k\_proj}, \texttt{v\_proj},
\texttt{o\_proj}, \texttt{gate\_proj}, \texttt{up\_proj},
and \texttt{down\_proj}.}
\end{table}

Unlike existing methods that train separate models for different target bitrates, this work aims to support image compression at multiple bitrates using a single model. Considering that frequency-domain coefficient distributions vary significantly across different quality factors $Q$ and become more complex at larger $Q$, we configure the training samples according to the quality-factor ranges and their associated prediction difficulty. Specifically, the candidate quality factors are divided into low-$Q$, medium-$Q$, and high-$Q$ groups, with sample proportions of 25\%, 25\%, and 50\%, respectively. This allocation covers the frequency-domain coefficient distributions associated with different quality factors while assigning a larger proportion to high-$Q$ samples, thereby strengthening the model’s capability to characterize coefficient distributions that are more difficult to predict. During two-stage fine-tuning, the first stage uses the first 400 images of the DIV2K training set, whereas the second stage uses the remaining 400 images. The datasets for both stages are constructed using the same data processing pipeline and quality-factor allocation scheme.

\subsubsection{Two-Stage Fine-Tuning} We conduct the fine-tuning experiments on NVIDIA GeForce RTX 4090 GPUs using LLaMA 3.2-1B as the base model. The main hyperparameter settings for the two-stage fine-tuning procedure are summarized in Table~\ref{tab:finetuning_settings}. The first stage focuses on learning the representations and output mappings of the newly introduced DCT tokens. Specifically, we freeze the model backbone and train only the token embedding and output projection layers, enabling the new tokens to adapt to the numerical characteristics and distribution patterns of frequency-domain coefficients. In the second stage, we initialize the model with the weights obtained from the first stage and freeze the token embedding layer, output projection layer, and original model parameters. We then apply LoRA~\cite{hu2022lora} to the model backbone for parameter-efficient fine-tuning, allowing the model to further learn the contextual dependencies among frequency-domain coefficients within each macroblock and improve its capability to predict DCT coefficient sequences.

\section{Experiments}

\subsection{Experimental Settings}
\subsubsection{Dataset}
Following \cite{li2024toward}, we evaluate the performance of FLM on the Kodak \cite{kodak1993kodak}, Tecnick, and CLIC2020 datasets. The Kodak dataset consists of 24 natural images with a resolution of $768 \times 512$, and the full-resolution images are used for inference. The Tecnick \cite{asuni2014testimages} and CLIC2020 \cite{toderici2020workshop} datasets contain 140 and 428 images, respectively. For these two datasets, we follow the preprocessing protocol in \cite{li2024toward}: each image is first resized such that its shorter side is 768 pixels and then center-cropped to a resolution of $768 \times 768$.
\subsubsection{Baseline}

To comprehensively evaluate the compression performance of FLM, we compare it with traditional image compression methods, including JPEG and JPEG-XL, as well as generative image compression methods, including DiffEIC~\cite{li2024toward}, Control-GIC~\cite{li2024once}, PerCo~\cite{careil2023towards}, Diff-ICMH~\cite{feng2026diff}, and OSCAR~\cite{guo2026oscar}. Given fixed transform and quantization results, FLM employs a large language model together with arithmetic coding to losslessly encode the quantized DCT coefficients. Therefore, FLM can also be regarded as a lossless JPEG recompression method. To evaluate its capability to compress frequency-domain coefficients in JPEG images, we further compare FLM with Lepton~\cite{horn2017design}, LLJPEG~\cite{sun2022lossless}, CMIX~\cite{Byroncmix}, and the method proposed by Xiang et al.~\cite{xiang2026efficient}. The results of PerCo and DiffEIC on the three test datasets were obtained from the DiffEIC repository, whereas the results for the other methods were obtained from our reproductions. For the lossless JPEG recompression task, the results of all compared methods were taken from the results reported in~\cite{xiang2026efficient}.

\subsubsection{Metric}
We employ peak signal-to-noise ratio (PSNR) to evaluate the objective quality of reconstructed images. To assess overall rate-distortion performance, we further report BD-Rate and BD-PSNR. For lossless JPEG recompression, since the evaluated methods preserve the original quantized DCT coefficients and reconstruction results, we compare their bits per pixel (BPP) under the same quality factor $Q$.

\subsection{Rate–Distortion Performance Comparison}

\subsubsection{Quantitative Comparison} 
\begin{figure*}[!htbp]
    \centering
    \includegraphics[width=1\textwidth]{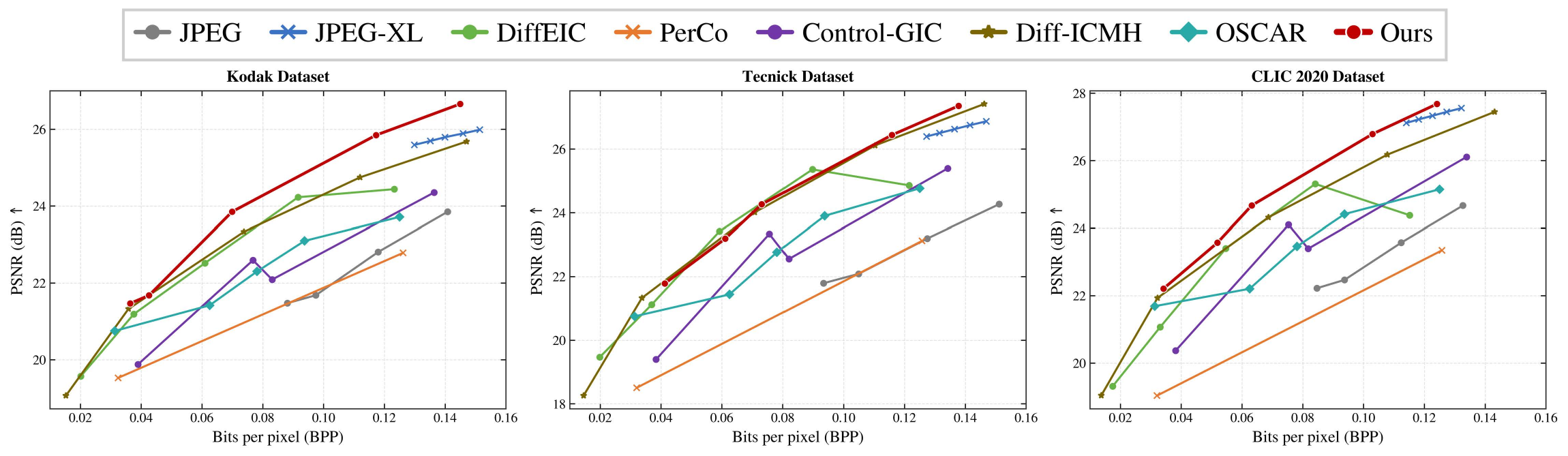}
    \caption{Rate-distortion performance comparison with existing methods on the Kodak, Tecnick, and CLIC2020 datasets.}
    \label{fig:rd_curve}
\end{figure*}

Fig.~\ref{fig:rd_curve} presents the PSNR-based R-D curves of different methods on the Kodak, Tecnick, and CLIC2020 datasets. On Kodak and CLIC2020, the R-D curve of FLM generally lies above and to the left of those of the competing methods, demonstrating its superior overall R-D performance across different bitrate settings. On Tecnick, FLM also achieves competitive performance, with results comparable to DiffEIC~\cite{li2024toward} and Diff-ICMH~\cite{feng2026diff}. The BD-Rate and BD-PSNR results further confirm these observations, as reported in Table~\ref{tab:bd_metrics}. Compared with JPEG, FLM reduces the BD-Rate by 52.82\%, 53.47\%, and 55.51\% on Kodak, Tecnick, and CLIC2020, respectively. These results indicate that FLM requires approximately half the bitrate of JPEG to achieve comparable reconstruction quality and that its performance improvements remain stable across different datasets.

\begin{table}[htbp]
\centering
\caption{Comparison of BD-PSNR and BD-Rate than JPEG baseline on Kodak, Tecnick, and CLIC2020 datasets. }
\label{tab:bd_metrics}
\renewcommand{\arraystretch}{1.3}
\setlength{\tabcolsep}{3.5pt}

\resizebox{\columnwidth}{!}{%
\begin{tabular}{lcccccc}
\toprule
\multirow{3}{*}{\textbf{Model}} &
\multicolumn{3}{c}{\textbf{BD-PSNR (dB) $\uparrow$}} &
\multicolumn{3}{c}{\textbf{BD-Rate $\downarrow$}} \\
\cmidrule(lr){2-4}
\cmidrule(lr){5-7}

& \textbf{Kodak}
& \textbf{Tecnick}
& \textbf{CLIC2020}
& \textbf{Kodak}
& \textbf{Tecnick}
& \textbf{CLIC2020} \\
\midrule

JPEG
& 0
& 0
& 0
& $0\%$
& $0\%$
& $0\%$ \\

DiffEIC~\cite{li2024toward}
& 2.24
& 3.14
& 2.49
& $-47.66\%$
& $-55.38\%$
& $-49.51\%$ \\

PerCo\textsuperscript{*}~\cite{careil2023towards}
& /
& /
& /
& /
& /
& / \\

Control-GIC~\cite{li2024once}
& 0.95
& 1.78
& 1.67
& $-32.17\%$
& $-42.30\%$
& $-40.49\%$ \\

Diff-ICMH~\cite{feng2026diff}
& 2.39
& 3.73
& 2.98
& $-50.14\%$
& $\mathbf{-55.92\%}$
& $-52.39\%$ \\

OSCAR~\cite{guo2026oscar}
& 1.26
& 2.15
& 1.74
& $-25.56\%$
& $-33.67\%$
& $-28.28\%$ \\

FLM (Ours)
& $\mathbf{3.30}$
& $\mathbf{3.83}$
& $\mathbf{3.80}$
& $\mathbf{-52.82\%}$
& $-53.47\%$
& $\mathbf{-55.51\%}$ \\

\bottomrule
\end{tabular}%
}
\vspace{2pt}
\begin{minipage}{\linewidth}
\footnotesize
\textsuperscript{*} PerCo fails to compute, as there are only 2 bitrate data shown as Fig.~\ref{fig:rd_curve}
\end{minipage}
\end{table}

\begin{figure}[!htbp]
    \centering
    \includegraphics[width=\columnwidth]{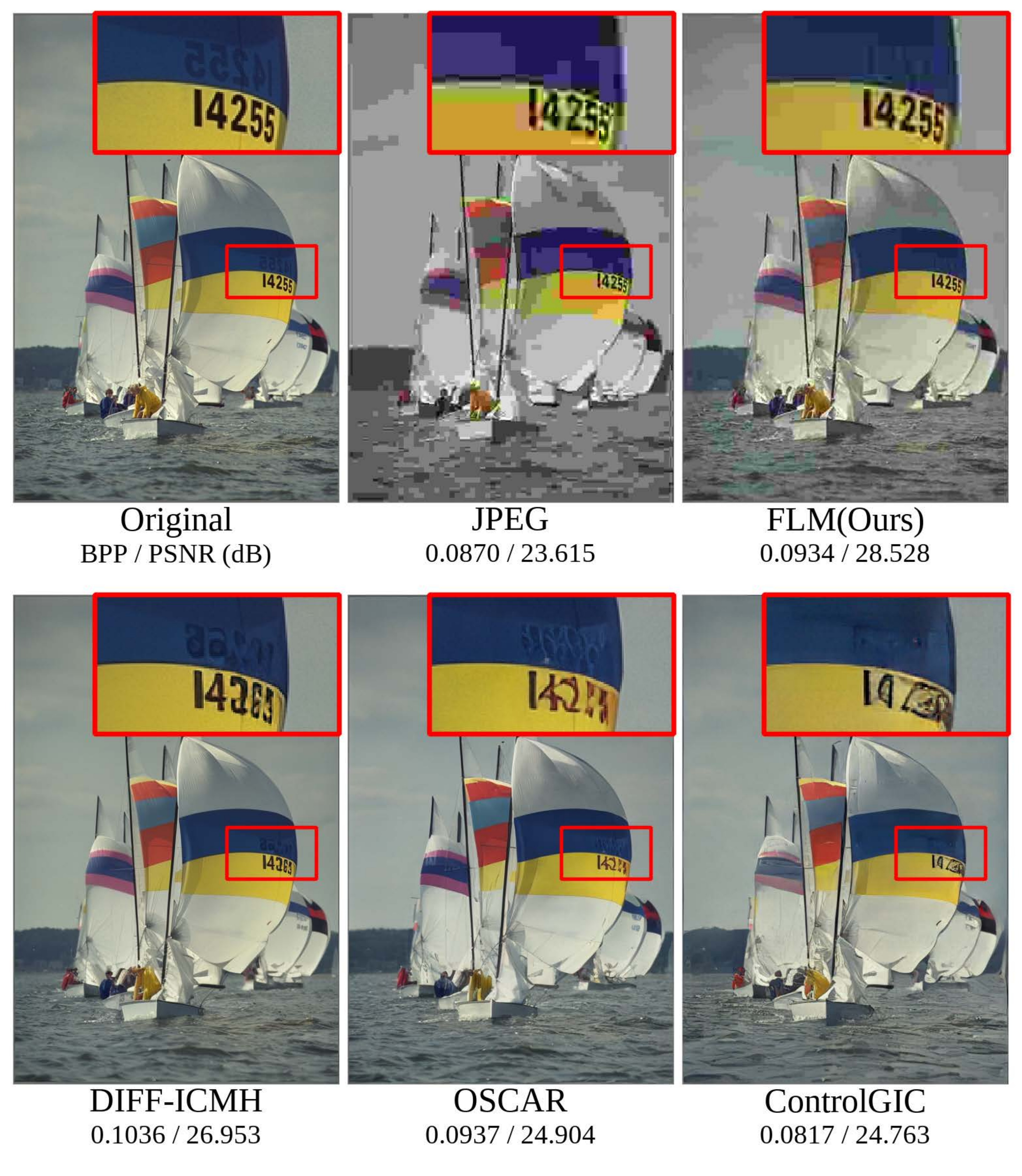}
    \caption{Subjective visual quality comparison of reconstructions produced by different methods for the Kodak image kodim09.}
    \label{fig:qualitative_analysis}
\end{figure}

\begin{table}[!htbp]
\centering
\caption{Comparison with JPEG lossless recompression methods in different quality factors.}
\label{tab:bpp_comparison}
\begin{tabular}{lcccc}
\toprule
\textbf{Method} &
\textbf{$Q=65$} &
\textbf{$Q=75$} &
\textbf{$Q=85$} &
\textbf{$Q=95$} \\
\midrule
JPEG \cite{wallace1991jpeg}
& 1.127 & 1.369 & 1.859 & 3.401 \\
JPEG-XL \cite{alakuijala2019jpeg}
& 0.960 & 1.173 & 1.595 & 2.849 \\
Lepton \cite{horn2017design}
& 0.896 & 1.102 & 1.520 & 2.786 \\
LLJPEG \cite{sun2022lossless}
& 0.894 & 1.107 & 1.536 & 2.853 \\
CMIX \cite{Byroncmix}
& 0.853 & 1.054 & 1.452 & 2.648 \\
Guo et al. \cite{guo2022practical}
& 0.778 & 0.965 & 1.341 & 2.500 \\
Guo et al.\textsuperscript{*} \cite{guo2022practical}
& 0.784 & 0.965 & 1.396 & 3.022 \\
Xiang et al. \cite{xiang2026efficient}
& \textbf{0.717} & \textbf{0.893} & \underline{1.255} & \underline{2.396} \\
FLM (Ours)
& \underline{0.723} & \underline{0.897} & \textbf{1.254} & \textbf{2.352} \\
\bottomrule
\end{tabular}

\vspace{2pt}
\begin{minipage}{\linewidth}
\footnotesize
\textsuperscript{*} denotes a variable-rate model.
\end{minipage}
\end{table}

\begin{figure}[!htbp]
    \centering
    \includegraphics[width=\columnwidth]{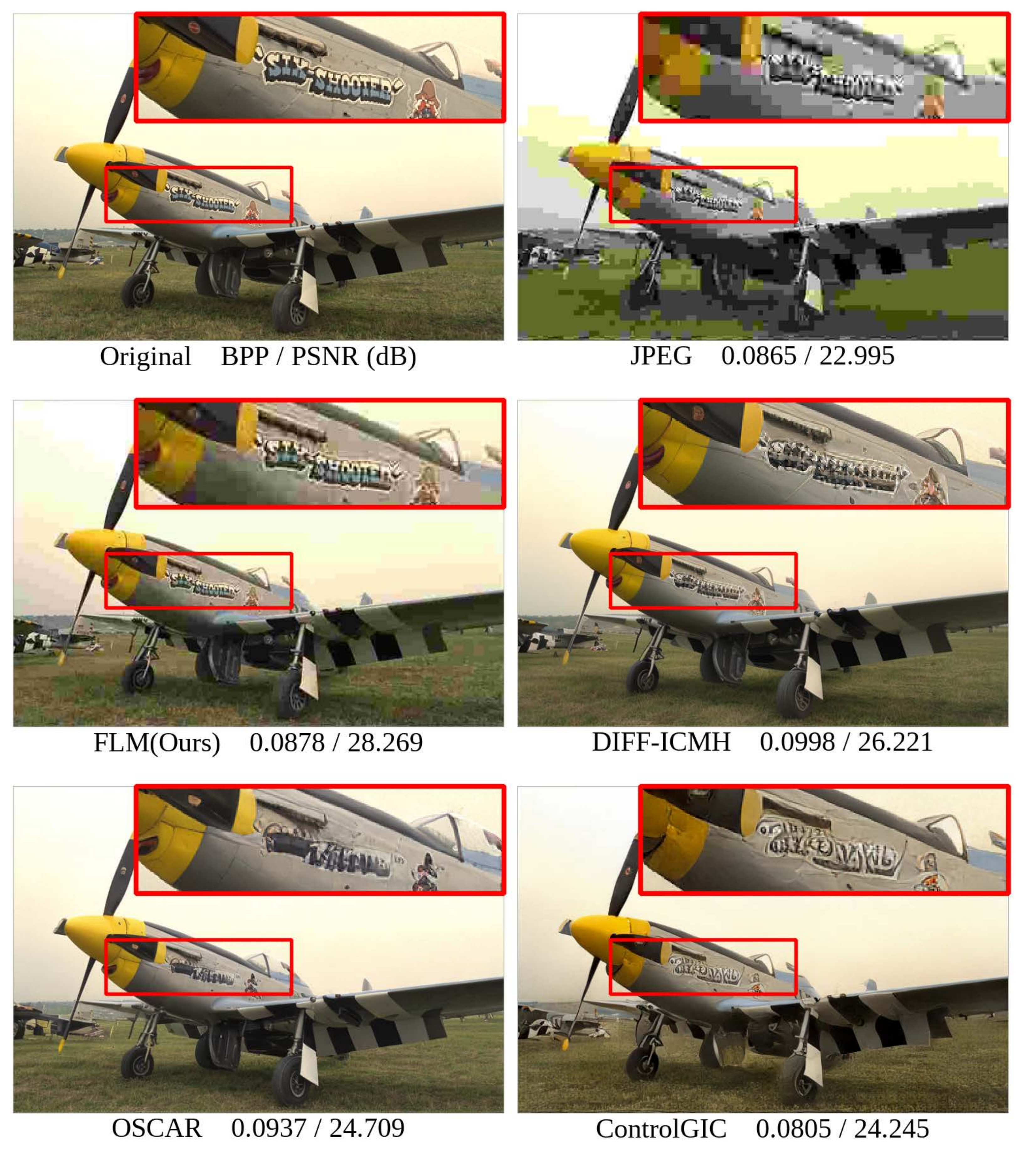}
    \caption{Subjective visual quality comparison of reconstructions produced by different methods for the Kodak image kodim20.}
    \label{fig:qualitative_analysis_2}
\end{figure}

\subsubsection{Qualitative Comparison} 
As shown in Fig.~\ref{fig:qualitative_analysis} and Fig.~\ref{fig:qualitative_analysis_2}, we qualitatively compare the reconstruction results of different methods and enlarge local details. At comparable BPP levels, our method achieves better semantic fidelity of fine details and more clearly recovers the numerical information on the sailboat. In contrast, the other methods suffer from blurred character boundaries and missing numerical semantics. However, in terms of overall subjective perceptual quality, our method is inferior to generative compression methods. One possible reason is that our method directly employs the standard JPEG quantization tables, while the quantization parameters are not jointly optimized during fine-tuning. Future work may introduce learnable quantization tables or end-to-end rate-distortion optimization to further improve the subjective visual quality of reconstructed images.

\subsection{Lossless Recompression Performance Comparison}

For lossless JPEG recompression, we compare different methods at quality factors $Q \in \{65, 75, 85, 95\}$. Since all evaluated methods preserve the original quantized DCT coefficients, they produce identical reconstruction results, and their compression performance can be directly compared in terms of BPP. As shown in Table.~\ref{tab:bpp_comparison}, FLM outperforms existing lossless JPEG recompression methods, including Lepton~\cite{horn2017design}, LLJPEG~\cite{sun2022lossless}, and CMIX~\cite{Byroncmix}, while achieving performance comparable to the method proposed by Xiang et al.~\cite{xiang2026efficient}. 

The consistent results across different quality factors indicate that FLM can effectively model DCT coefficient distributions with varying sparsity and numerical ranges. These results further demonstrate the advantages of employing a large language model as an entropy model to capture contextual dependencies and predict the conditional probability distributions of image frequency-domain coefficients.

\begin{table*}[!htbp]
\centering
\caption{Ablation studies on fine-tuning strategies, macroblock sizes, and functional module designs.}
\label{tab:ablation_study}
\renewcommand{\arraystretch}{1.12}
\setlength{\tabcolsep}{3.2pt}

\resizebox{\textwidth}{!}{%
\begin{tabular}{
    >{\centering\arraybackslash}p{1.7cm}
    >{\centering\arraybackslash}p{3.4cm}
    *{8}{c}
    c
}
\toprule
\multirow{4}{1.7cm}{\centering\textbf{Experiment}} &
\multirow{4}{3.4cm}{\centering\textbf{Model / Setting}} &
\multicolumn{8}{c}{\textbf{Quality Factor}} &
\multirow{3}{*}{\textbf{$\Delta$BPP$_{\mathrm{avg}}$}} \\
\cmidrule(lr){3-10}

& & \multicolumn{2}{c}{\textbf{5}}
  & \multicolumn{2}{c}{\textbf{35}}
  & \multicolumn{2}{c}{\textbf{55}}
  & \multicolumn{2}{c}{\textbf{75}} & \\

\cmidrule(lr){3-4}
\cmidrule(lr){5-6}
\cmidrule(lr){7-8}
\cmidrule(lr){9-10}

& & \textbf{BPP} & \textbf{$\Delta$BPP}
  & \textbf{BPP} & \textbf{$\Delta$BPP}
  & \textbf{BPP} & \textbf{$\Delta$BPP}
  & \textbf{BPP} & \textbf{$\Delta$BPP} & \\
\midrule

\multirow{3}{1.7cm}{\centering Fine-tuning}
& Baseline 
& 0.194 & --
& 1.008 & --
& 1.359 & --
& 1.945 & --
& -- \\
\cmidrule(lr){2-11}

& Fine-tune
& 0.070 & $-64.026\%$
& 0.433 & $-57.054\%$
& 0.605 & $-55.461\%$
& 0.897 & $-53.872\%$
& $-57.603\%$ \\

\midrule

\multirow{4}{1.7cm}{\centering Macroblock size}
& Baseline (\shortstack{$16\times8$})
& 0.194 & --
& 1.008 & --
& 1.359 & --
& 1.945 & --
& -- \\
\cmidrule(lr){2-11}

& \shortstack{$8\times4$}
& 0.204 & $+5.007\%$
& 1.054 & $+4.624\%$
& 1.431 & $+5.297\%$
& 2.058 & $+5.796\%$
& $+5.181\%$ \\
\cmidrule(lr){2-11}

& \shortstack{$1\times1$}
& 0.537 & $+176.447\%$
& 1.765 & $+75.145\%$
& 2.252 & $+65.684\%$
& 3.049 & $+56.770\%$
& $+93.512\%$ \\

\midrule

\multirow{4}{1.7cm}{\centering
Module design}
& Baseline
& 0.194 & --
& 1.008 & --
& 1.359 & --
& 1.945 & --
& -- \\
\cmidrule(lr){2-11}

& \shortstack{w/o Prompt}
& 0.212 & $+9.021\%$
& 1.025 & $+1.733\%$
& 1.381 & $+1.639\%$
& 1.964 & $+0.957\%$
& $+3.338\%$ \\
\cmidrule(lr){2-11}

& \shortstack{w/o Range Constraint}
& 0.311 & $+59.917\%$
& 1.263 & $+25.324\%$
& 1.659 & $+22.067\%$
& 2.316 & $+19.052\%$
& $+31.590\%$ \\

\bottomrule
\end{tabular}%
}

\vspace{2pt}

\begin{minipage}{\textwidth}
\footnotesize
\raggedright
\textit{Note:} $\Delta\mathrm{BPP}$ are computed from unrounded BPP measurements.
\end{minipage}

\end{table*}

\subsection{Ablation Study}
Table~\ref{tab:ablation_study} presents the ablation results of FLM on the Kodak dataset, examining the effects of the fine-tuning strategy, macroblock size, and functional modules on coding performance at quality factors $Q\in\{5,35,55,75\}$. 
To quantify the relative change in coding rate with respect to the corresponding baseline, we define
\begin{equation}
\Delta\mathrm{BPP}(Q)
=
\frac{
\mathrm{BPP}_{\mathrm{cfg}}(Q)
-
\mathrm{BPP}_{\mathrm{base}}(Q)
}{
\mathrm{BPP}_{\mathrm{base}}(Q)
}
\times 100\%,
\label{eq:relative_bpp_change}
\end{equation}
where $\mathrm{BPP}_{\mathrm{cfg}}(Q)$ and
$\mathrm{BPP}_{\mathrm{base}}(Q)$ denote the BPP values of the evaluated configuration and baseline at quality factor $Q$, respectively. 

Except for the fine-tuned experiment, all configurations use the original weights of the vocabulary-expanded model. The non-fine-tuned model with a macroblock size of $16\times8$ is adopted as the common baseline and repeated in the first row of each experimental group for consistent comparison and analysis. Here, a macroblock size of $H\times W$ denotes that the coefficient sub-blocks are arranged in $H$ rows and $W$ columns. Prompt and Range Constraint are described in Secs.~\ref{3_2} and~\ref{3_3}, respectively.
\subsubsection{Fine-Tuning}
After fine-tuning, FLM achieves substantial BPP reductions across all quality factors, with an average bitrate saving of \(57.603\%\). This result indicates that fine-tuning enables the model to better adapt to the statistical distributions of DCT coefficients and improves the probability estimation of target tokens, thereby enhancing entropy coding efficiency.
\subsubsection{Macroblock Size}
Compared with the \(16\times8\) baseline, the \(8\times4\) and \(1\times1\) macroblocks incur average bitrate overheads of \(5.181\%\) and \(93.512\%\), respectively. This result suggests that, among the evaluated configurations, a larger macroblock provides richer spatial context and facilitates the modeling of dependencies among adjacent coefficient sub-blocks.
\subsubsection{Module Design}
Removing the prompt introduces an average bitrate penalty of \(3.338\%\), demonstrating that information such as the channel type, quality factor, and coefficient ranges contributes to conditional probability prediction. Removing Range Constraint results in an average bitrate penalty of \(31.590\%\), indicating that this module effectively excludes invalid candidate tokens and prevents them from consuming probability mass, thereby improving entropy coding efficiency.

\section{Conclusion}
This paper proposes FLM, a large language model (LLM)-based lossy generative image compression method. It transforms images into the frequency domain and generates bitstreams through macroblock-based DCT coefficient tokenization method, LLM-based next-coefficient prediction, and arithmetic coding. At the decoder, the DCT coefficient sequences are autoregressively recovered to reconstruct the images. To improve compression efficiency, we construct a frequency-domain coefficient prediction dataset and adopt a two-stage fine-tuning strategy. Experimental results demonstrate that FLM outperforms conventional codecs, including JPEG and JPEG-XL, as well as generative compression methods such as Control-GIC and OSCAR, on multiple datasets. 
Since the large language model predicts and decodes frequency-domain coefficients in an autoregressive manner, it suffers from slow computation and high computational and memory requirements, making the proposed method difficult to deploy in real-time image transmission scenarios. Future work will focus on lightweight architecture design and inference acceleration to reduce computational complexity and resource consumption, thereby further improving runtime efficiency.

\bibliographystyle{IEEEtran}
\bibliography{refs}

\end{document}